\documentclass{article}

\pdfoutput=1

\usepackage{arxiv}

\usepackage[utf8]{inputenc} 
\usepackage[T1]{fontenc}    
\usepackage{hyperref}       
\usepackage{url}            
\usepackage{booktabs}       
\usepackage{amsfonts}       
\usepackage{nicefrac}       
\usepackage{microtype}      

\usepackage{graphicx} 
\graphicspath{{images/}}
\usepackage{bm}
\def\x{\bm x}
\def\y{\bm y}
\def\s{\bm s}
\usepackage{amsmath} 
\title{Defending Against Adversarial Attacks
by Suppressing the Largest Eigenvalue of Fisher Information Matrix}

\author{
  Chaomin Shen
  \\
  School of Computer Science and Technology\\
  East China Normal University\\
  Shanghai, China \\
  \texttt{cmshen@cs.ecnu.edu.cn} \\
   \And
 Yaxin Peng \\
  Department of Mathematics\\
  Shanghai University\\
  Shanghai, China \\
  \texttt{yaxin.peng@shu.edu.cn} \\
  \AND
   Guixu Zhang\\
  School of Computer Science and Technology\\
  East China Normal University\\
  Shanghai, China \\
  \texttt{gxzhang@cs.ecnu.edu.cn} \\
  \And
  Jinsong Fan\thanks{Corresponding author}\\
  College of Mathematics and Physics\\
  Wenzhou University \\
  Wenzhou, Zhejiang, China \\
  \texttt{fjs@wzu.edu.cn}
}

\date{}

\begin{document}
\maketitle

\begin{abstract}
We propose a scheme for defending against
adversarial attacks by suppressing
the largest eigenvalue of the Fisher information matrix (FIM).
Our starting point is one explanation on the rationale of adversarial examples. Based
on the idea of the difference between a benign sample and its adversarial example
is measured by the Euclidean norm, while the difference between their classification
probability densities
at the last (softmax) layer
of the network could be measured by the Kullback-Leibler (KL) divergence,
the explanation shows that the
output difference is a quadratic form of the
input difference.
If the eigenvalue of this quadratic form (a.k.a. FIM) is large,
the output difference becomes large even when the input difference is small,
which explains the adversarial phenomenon.
This makes the adversarial defense possible by controlling the eigenvalues
of the FIM. Our solution is
adding one term representing the trace of the FIM to the loss function
of the original network,
as the largest eigenvalue is bounded
by the trace. Our defensive scheme is verified by experiments
using a variety of common attacking methods on
typical deep neural networks, e.g. LeNet, VGG and ResNet,
with datasets MNIST, CIFAR-10, and German Traffic Sign Recognition Benchmark (GTSRB).
Our new network, after adopting the novel loss function and retraining, has
an effective and robust defensive capability,
as  it decreases the fooling ratio of the generated adversarial examples,
and remains the classification accuracy of the original network.

\end{abstract}


\section{Introduction}
Adversarial examples are samples added with carefully designed perturbations,
such that those perturbed samples will be misclassified by the Deep
Neural Networks (DNNs) for classification
\cite{szegedy2013intriguing,Biggio}. In certain critical
circumstances, such as autonomous driving or security sensitive tasks, it is crucial to avoid
such adversarial phenomenon. Therefore, defending against adversarial attacks becomes a hot topic in machine learning
\cite{AML_book}.

In order to avoid/alleviate the adversarial phenomenon, it is important to reveal the rationale of adversarial examples.
Many explanations have been provided, yet no consensus has been reached.
For example, \cite{szegedy2013intriguing} suggested that
the adversarial phenomenon
is due to the excessive non-linearity of the neural networks. Later the idea was modified by  models being too linear \cite{Goodfellow2014Explaining}.
Another explanation claimed that
the phenomenon results from the high curvature regions on the decision boundary
\cite{moosavi2017analysis}.

In \cite{our_AAAI}, the authors suggested that the vulnerability of DNNs may be
caused by the large value of the largest eigenvalue of the Fisher Information Matrix (FIM)
induced by the input sample.
Their idea is that the input
difference of the DNNs between a benign sample and its adversarial example
is measured by the Euclidean norm, while the output difference between their classification
probability density vectors at the last (softmax) layer
could no longer be measured by the Euclidean norm, and should be
measured by other suitable ``distance'' such as the Kullback-Leibler (KL) divergence.
Then, the output difference is a quadratic form of the
input difference, and the quadratic matrix is the FIM.
Thus, the adversarial example can be constructed by setting the perturbation
direction as the direction of the eigenvector for the largest eigenvalue.
This method is called
One Step Spectral Attack (OSSA).

\begin{figure}
\centering
\includegraphics[width=0.5\textwidth]{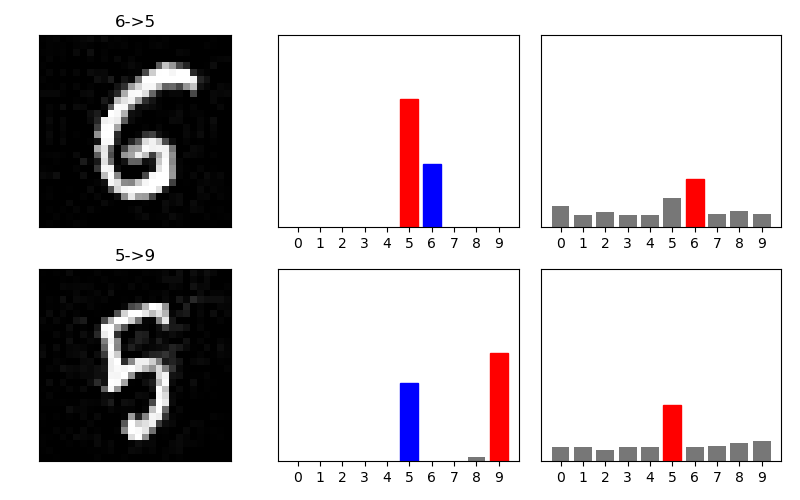}
\caption{Visualization for the results of our defensive scheme.
Left column: two adversarial examples generated on MNIST via the attack method OSSA.
Middle column: the predicted label distributions by the original DNN,
i.e., result without using our defensive scheme. The blue/red bar represents the true/misclassified label, respectively. In row 1, '6' is misclassified as '5', and in row 2, '5' is misclassified as '9'.
Right column: the predicted label distribution of our scheme by modifying the loss function of the original DNN. This figure demonstrates that we survive under the adversarial
attack and obtain the correct result. With our scheme, the label distribution is smoothed but the
correct label has the highest probability.}
\label{fig_distributions}
\end{figure}

Inspired by \cite{our_AAAI}, in this paper we propose a method for defending against adversarial
attacks by only modifying the loss function of the original DNN
so that the largest eigenvalue of the FIM is suppressed.
Here we first illustrate our result in Figure  \ref{fig_distributions}.
Details will be given in Section \ref{proposed}.

The novelties of the paper lie in:

1) Our scheme is simple yet effective, as it only
adds one regularization term to the  loss function of original DNNs and needs to retrain only once.
With our scheme, the capability of defending against adversarial examples can be significantly improved;

2) Our scheme is explainable, as it is based on mathematical deduction.
After the softmax layer, its output probability distribution is similar
to that of label smoothing \cite{LSR},
while ours provides a reasonable explanation.

The rest of the paper is organized as follows.
Section \ref{Related_work}
reviews some related work on adversarial attacks and defense.
Section \ref{proposed} describes our defensive scheme in detail with mathematical deduction,
and makes a comparison with label smoothing.
Section \ref{experiments} shows the experiments on defending against various white-box and black-box
attacks on a variety of datasets and DNNs. The paper is concluded in
Section \ref{conclusion}.

\section{Related Work}
\label{Related_work}
In this section, we review some related work on adversarial attacks and defense.

\subsubsection{Adversarial attacks }

Adversarial attacks can be classified based on different criteria,
while the most important and common criterion is according to
the measure of the perturbation, i.e., they can be classified as
$l_0$, $l_2$ and $l_\infty$ attacks. Adversarial attacks can also
be  targeted or non-targeted according to the misclassified
label concerned, one-step attack or multi-step attack according to
whether the adversarial example is generalized by one-step or by iteration,
or white-box or black-box attack according to whether the
network structure and corresponding weights are known.

\cite{Goodfellow2014Explaining} proposed a simple and fast
$l_\infty$ attack, called the Fast Gradient Sign Method
(FGSM), by  using
the sign of the gradient to generate perturbation.
\cite{Kurakin2016Physical,kurakin2016adversarial}
extended the gradient method to $l_2$ norm, and proposed
the Basic Iterative Method (BIM), and other consequent more powerful
attacking methods.
\cite{Carlini2016Towards} proposed the Carlini-Wagner (CW) attack, which
includes $l_0$, $l_2$, $l_\infty$ norms and can be applied to targeted or non-targeted
attacks.
\cite{Moosavi-Dezfooli} developed
DeepFool, an $l_2$-norm and non-targeted method which uses iteration to push the image to the
classification boundary iteratively.
Jacobian-based Saliency Map Attack (JSMA),
an $l_0$-norm targeted method proposed by
\cite{papernot},
minimizes the number of modified pixels
so that the image is misclassified as a particular (wrong) target class.
Those more powerful methods all adopt the iterative optimization methods, such
that the perturbation is small yet effective.
To achieve such a goal, simple methods, e.g., FGSM, usually need to
conduct searches on certain particular perturbation space, such as binary search.

\subsubsection{Adversarial defenses}
The adversarial phenomenon naturally spawns
many works on defending against adversarial attacks for DNNs.
To date, these adversarial defenses can be roughly categorized into the following three types:

1) Data pre-processing. This kind of method pre-processes the input data during the
training and testing, which includes
image gradient processing, transforming, denoising,  enhancing,
or  a small scale network training to pre-process the input.

2) Classification model modification. This kind of method increases/deletes/modifies certain
layers, or modifies the loss function and the activation function.

3) Robust learning. This kind of method includes adversarial training
and robust gradient descent techniques. Note that here adversarial training
is in its narrow sense, meaning using adversarial examples as a part of the training data.

Adversarial training \cite{szegedy2013intriguing} trains the model by repeatedly feeding the
adversarial examples into the training set, such that the new model can possess certain defensive
capability.
Due to its simplicity and effectiveness, many adversarial attacking methods, e.g., \cite{our_AAAI}, also used adversarial training for evaluation. \cite{our_AAAI} also
proposed a novel approach for adversarial detection, i.e., when an adversarial example is
detected, warning is triggered and further processing aborts.

\cite{madry} proposed a robust optimization method called Projected Gradient Descent (PGD).
Based on the corollary of a classical result in \cite{Danskin}, PGD uses
the gradient descent method
to obtain an optimized defensive network. Note that PGD is also an attack method
from another point of view. \cite{madry} was adopted by many defensive methods such as
the thermometer encoding \cite{Buckman}.

More recently, the method of Label Smoothing Regularization (LSR) receives attention in adversarial
defense. Introduced by \cite{LSR}, LSR was first used for improving the performance of
Inception network and now has become a common technique for DNN regularization. LSR can improve the generalization capability and also increase the classification
accuracy by a small amount.
The role of LSR in adversarial defense was observed by experiments in
\cite{Warde}.
\cite{Goibert} proposed many variants of LSR for adversarial defense.

\section{Proposed Scheme}
\label{proposed}
In this section, we first summarize the mechanism of adversarial phenomenon
proposed in \cite{our_AAAI}. Then based on that mechanism, we propose our
solution for defending against adversarial attacks,
followed by the analysis of our scheme.

\subsection{Our starting point}
\label{Preliminary}
The starting work of our defensive scheme is \cite{our_AAAI}.
Its basic idea can be summarized as follows.

Given a (gray) image of $m\times n$, the aim of a classification problem is
to classify it as $K$ classes.

Suppose that the image has been pulled as
a (column) vector $\bm x$ with the length $m\times n$, and its perturbed image is
$\x'=\x+\bm \eta$, where $\bm\eta$ is the perturbation.
After certain DNN, their softmax layer outputs are $\s(\x)=[p_1(\x),\cdots,p_K(\x)]^T$
and $\s(\x+\bm\eta)=[p_1(\x'),\cdots,p_K(\x')]^T$,
where $p_i\ge 0$ for $i=1,\dots,K$ and $\sum_{i=1}^K p_i=1$.

Denote the probability that
$\x$ belongs to the $i$-th class by $p(\y_i|\x)$,
where $\y_i$ is a one hot vector $\y_i=[y_1,\cdots,y_i,\cdots, y_K]^T$ where
$y_i=1$ and $y_j=0$ for $j\ne i$.
Thus, $p(\bm y_i|\x)=p_i(\x)$, and
$\s(\x)=[p(\y_1|\x),\cdots,p(\y_K|\x)]^T$.
The classification label $i$  is
determined by $i=\arg\max\limits_j p_j(\bm x)$.

The distance between $\x$ and $\x+\bm\eta$ can be measured by $l_k$ norm, where
$k$ can be chosen as 1, 2, or $\infty$. In \cite{our_AAAI}, $k$ is chosen as 2,
i.e., the Euclidean
norm is used. However, the Euclidean norm cannot be used as a measure of distance between
$\s(\x)$ and $\s(\x')$, as $\{\s(\x)\}$ does not form a linear space.
A suitable measure is the Kullback-Leibler (KL) divergence, denoted by $D_{KL}$.

Let $\y$ be the random variable ranging from $\y_1$ to $\y_K$
with the distribution density $p(\y|\x)$,
where $p(\y|\x)=p(\y_i|\x)=p_i(\x)$ if $\y=\y_i$ for $i=1,\cdots,K$. Then we
can expand $D_{KL}$ using the 2nd order Taylor expansion, i.e.,
\begin{equation}
 D_{KL}(\s(\bm{x})\|\s(\bm{x}+\bm{\eta}))
 =\mathbb{E}_{\bm y}[\log\frac{{p(\y|\bm{x})}}{p(\y|\bm{x}+\bm{\eta})}]
  \approx\frac{1}{2}\bm{\eta}^{T}\bm{G}_{\bm{x}}\bm{\eta},
\label{quadratic}
\end{equation}
where $\bm{G}_{\bm{x}}=\mathbb{E}_{\y}[(\nabla_{\bm{x}}\log{p(\y|\bm{x})})(\nabla_{\bm{x}}
\log{p(\y|\bm{x})})^{T}]$
is the FIM of $\bm{x}$.

Note that the quadratic form
(\ref{quadratic}) becomes large if the largest eigenvalue of the FIM,
$\lambda_{\max}(G_{\x})$, is large.
Therefore, adversarial phenomenon may occur if  $\lambda_{\max}(G_{\x})$ is large.

The quadratic form (\ref{quadratic}) can be used to generate the adversarial attack.
OSSA is constructed by maximizing (\ref{quadratic}) under some constraints,
i.e.,
$$
\label{objective}
  \max_{\bm{\eta}}\bm{\eta}^{T}\bm{G}_{\bm{x}}\bm{\eta} \quad{}
  \mathrm{s.t.}\ \|\bm{\eta}\|_{2}^{2}=\varepsilon,
  \ J(\y,\bm{x}+\bm{\eta})> J(\y,\bm{x}),
$$
where $\varepsilon$ denotes the squared norm of the
perturbation,  and $J$ is the loss function.

Therefore, the problem of constructing adversarial examples
is converted to the problem of finding the largest eigenvalue $\lambda_{\max}(G_{\x})$
and its corresponding eigenvector $\bm\eta$
of $G_{\bm x}$, i.e., $G_{\bm x}\bm\eta=\lambda_{\max}\bm \eta$.
That is, the perturbation $\bm\eta$ should be the product of the length $\sqrt\varepsilon$ and
the normalized eigenvector
corresponding to the largest eigenvalue.

\subsection{Our Scheme}
\label{our_approach}
This subsection is devoted to how to construct our defensive scheme for adversarial attacks.

Since larger eigenvalue of $G_{\bm x}$ may cause
the larger difference of KL divergence, and consequently cause the adversarial phenomenon,
one solution for defending against
adversarial attacks is to control $\lambda_{\max}(G_{\x})$
induced by the input sample.

A natural approach is to add a regularization term to the loss function of the original network
to suppress the largest eigenvalue
$\lambda_{\max}(G_{\bm x})$ of the FIM $G_{\x}$, i.e., modify the loss function as
\begin{equation}
L(\Theta)+\mu \cdot \lambda_{\max}(G_{\bm x}),
\label{loss_lambda_max}
\end{equation}
where $\mu$ is the regularization parameter,
$L(\Theta)$ is the loss function of the original network, and
$\Theta$  is the set of parameters of the original network.

Two problems prevent the direct use of the loss function
(\ref{loss_lambda_max}).
One is that the matrix $G_{\bm x}$ is too large.
For example, given a $1000\times 1000$ image, it is  a vector of length $10^6$
after conversion, which means that $G_{\bm x}$ is $10^6\times 10^6$.
The other is that
it is difficult to explicitly write down the formula of
$\lambda_{\max}(G_{\bm x})$, even if we can tackle the big matrix.

The first problem can be settled by the following strategy.
We turn our focus from $G_{\x}$ to a new matrix $G_{\s}$,
where $\s$ is the output of the softmax layer
$\bm s=[p_1(\bm x), \cdots,  p_K(\bm x)]^T$.
Similar to $G_{\x}$, we have
$$
G_{\bm s}=
\mathbb E_{\y}[\nabla_{\bm s}\log p(\bm y|\bm s)\cdot [\nabla_{\bm s}\log p(\y|\bm s)]^T].
$$
Note that $G_{\bm s}$ is a $K\times K$ positive definite matrix.

We have
\begin{eqnarray*}
\bm\eta^T G_{\bm x}\bm\eta
    &=& \bm\eta^T \mathbb E_{\y}[J^T\nabla_{\bm s}\log p(\y|\bm s)
        [J^T\nabla_{\bm s}\log p(\y|\bm s)]^T]\bm \eta\\
    &=& \bm\eta^T J^T \mathbb E_{\y}[\nabla_{\bm s}\log p(\y|\bm s)\cdot
        [\nabla_{\bm s}\log p(\y|\bm s)]^T]J\bm \eta,
\end{eqnarray*}
where $J=\Big(\frac{\partial s^i}{\partial x^\alpha}\Big)$ is a $K\times mn$ Jacobian of $\bm s=\bm s(\bm x)$,
and $\nabla_{\bm x}=J^T\nabla_{\bm s}$.

Then,
$$
\bm\eta^T G_{\bm x}\bm\eta
    = \bm \eta^T J^T  G_{\bm s}J\bm\eta.
$$

Thus the first problem is settled by converting calculating the largest eigenvalue
and corresponding eigenvector of a large $mn\times mn$ matrix into calculating those of a
much smaller $K\times K$
matrix.

Then the loss function can be further written as
\begin{equation}
L(\Theta)+\mu \lambda_{\max} (G_{\bm s}).
\label{G_s}
\end{equation}

Note that although (\ref{loss_lambda_max}) is greatly simplified as (\ref{G_s}),
the problem of no explicit expression for $\lambda_{\max} (G_{\bm s})$ still
exists.

The second problem could be solved via replacing the
largest eigenvalue $\lambda(G_{\s})$ by the trace of $G_{\bm s}$, as the
trace equals the summation of all eigenvalues which are all positive due to
the positive definiteness of $G_{\bm s}$.
Thus, our loss function changes to
$$
L(\Theta)+\mu\operatorname{tr}G_{\bm s}.
$$

The trace of $G_{\bm s}$ can be calculated as follows.
\begin{eqnarray*}
\operatorname{tr}G_{\s}
    &=& \operatorname{tr} \mathbb E_{\bm y}[\nabla_{\bm s} \log p(\bm y|\bm s)\cdot
        [\nabla_{\bm s}\log p(\bm y|\bm s)]^T]\\
    &=& \int_{\bm y}p(\bm y|\bm s)[\operatorname{tr}((\nabla_{\bm s}\log p(\bm y|\bm s))^T
        (\nabla_{\bm s}\log p(\bm y|\bm s))]\\
    &=& \int_{\bm y}p(\bm y|\bm s)\cdot \| \nabla_{\bm s}\log p(\bm y|\bm s)\|_2^2\\
    &=& \sum_{i=1}^K p_i\sum_{j=1}^K (\nabla_{p_j}\log p(\y_i|\bm s))^2\\
    &=& \sum_{i=1}^K p_i\sum_{j=1}^K (\nabla_{p_j}\log p_i)^2\\
    &=&\sum_{i=1}^K \frac 1{p_i}.
\end{eqnarray*}
Thus
the final loss function for the defensive scheme turns to be

\begin{equation}
\widetilde{L}(\Theta) = L(\Theta)+\mu \cdot \sum_{i=1}^K \frac 1{p_i}\quad
\text{s.t.\ }\sum_{i=1}^K p_i=1.
\label{final_loss}
\end{equation}

To summarize, we improve the defensive capability of the
original DNNs by modifying its loss function as (\ref{final_loss}),
and keep everything else. Of course, due to this modification, the new model
should be retrained.  That is, the optimal parameters for our defensive network can be obtained by solving
$$
\Theta^*=\arg\min_{\Theta} \widetilde{L}(\Theta).
$$

\subsection{Analysis of the proposed scheme}
Note that the solution of $\arg\min\limits_{p_k}\sum_{i=1}^K\frac 1{p_i}$ is
$p_1=\cdots=p_k=\frac 1K$ under the constraints,
which indicates that the effect of the regularization term
$\sum_{i=1}^K\frac1{p_i}$ is to force
$[p_1,\cdots, p_K]^T$ move towards the central point $[1/K,\cdots, 1/K]^T$.
In other words, it will prevent from the point to the positions such as
$[0,\cdots,1,\cdots,0]^T$.
It is natural to worry that this will decrease the classification accuracy
of DNNs. In what follows, we claim that this is not the case.

Adding the
term $\sum_{i=1}^K\frac1{p_i}$
indeed will cause the minimizer of $[p_1,\cdots,p_K]^T$ move towards $[1/K,\cdots,1/K]^T$,
but that point will not be reached due to the existence of the first term in the loss function
(\ref{final_loss}). This tendency of moving towards the center part of the simplex is no harm, as
what we really care is the correctness of  $i^*=\arg\max\limits_i p_i$, not the value of $p_{i^*}$.
Therefore, we do not pursue the large value of $p_{i^*}$, as long as $p_{i^*}$ reaches maximum among
all $p_i$ for $i=1,\cdots, K$.

The above argument also reveals a fact that, although it may be different from intuition, high
confidence (or over-confidence) on the classification result of one sample is unreliable sometimes.
The reason is that over-confidence on one Class $i$ means
that the
value of $p_i$ is large and consequently some values of $p_j$ should be small for $j\ne i$.
One extreme case is some $p_j =0$ and therefore $\sum_{i=1}^K\frac 1{p_K}\to \infty$.
In other words, this over-confidence sample is sensitive
to the adversarial perturbation and therefore vulnerable to adversarial attacks.

The experiments provided in Section \ref{experiments} will demonstrate that, 
after using our loss function,
the classification accuracy remains while the risk of adversarial phenomenon decreases.

\subsection{Comparison with Label Smoothing Regularization}
Both our scheme and LSR tend to smooth the labels, while the desired label still has the
highest probability.
To emphasis their difference, we
explain LSR and compare these two methods.

The basic procedure of LSR can be described as two steps.
Firstly, for $k\in\{1,\cdots, K\}$,  modify the
component of one-hot label as:
$$
y_k^{LSR}=y_k(1-\alpha)+\alpha/K,
$$
where $\alpha\in (0,1)$ is a hyperparameter,
$y_k$ is the label component whose
value is 1 for the correct class and 0 for the rest,
and $K$ is the total number of classes.
Second, retrain the network with the new labels.

LSR makes the classification clusters much tighter, according to the experiments in
\cite{Muller_et_al}.
This, to some extent, explains why LSR is effective in adversarial defense.
The underlying rationale of LSR, however, is still unknown. LSR needs to have
a not commonly accepted
prior assumption, i.e., the labels are uniformly distributed
for classes other than  the true class, while
our scheme is based on strict mathematical deduction
and does not have such a premise. Therefore, ours has a better explainability.

\section{Experiments}
\label{experiments}
The main purpose of our experiments
is to demonstrate the effectiveness
and robustness of our defensive scheme,
and therefore to demonstrate the correctness of using the loss function (\ref{final_loss}).

How to evaluate
the capability of a network for defending against adversarial examples  is a complicated issue.
Here, we illustrate four aspects of the complexities.

1) Adversarial examples is a relative concept.
For instance, a sample is an adversarial example for Network $A$,
it may be a benign sample for Network $B$, and vice versa.
For instance, the perturbed digits '5' and '6'
in Figure \ref{fig_distributions} are adversarial examples
for the network corresponding to Column 2, but not for Column 3.

2) The effect of the perturbation norm. If the norm is sufficiently large,
the perturbed image will be significantly different from the original image, therefore it
will be ``misclassified'' as another class. In other words, every sample will be an adversarial
example if its perturbation norm is sufficiently large. Therefore, when talking about how to defend
against adversarial attack, we usually assume that the perturbation is small. However,
the large perturbation norm can be used to show the defensive capability. Given a sample with large
perturbation, if it is an adversarial example in Network $A$ but not in Network $B$, we can say that Network $B$ is ``better'' for this particular sample.

3) It is not meaningful for talking about defensive capability for a single sample.
So some statistical index should be used. Say, given 1000 perturbed images, $a\%$ of them
are misclassified in Network $A$, and $b\%$ of them are misclassified in Network $B$.
If $b<a$, then we can conclude $B$ is better.

4) Defensive capability depends on different attacking methods.
A network, which can defend against perturbed examples generated from the
attacking method 1, may fail for the attacking method 2,
and vice versa.

Here we design the experiments according to four aspects mentioned above.

For 1), in what follows,
our scheme means for a given original network, using the loss function (4) and we have re-trained
the network and have obtained its new parameters $\Theta^*$. Using the term in the previous
description, the original network is $A$ and our scheme is $B$.

For 2), in the case of $l_2$ norm, we use the perturbation norm defined as
$\varepsilon=\sqrt{\sum_{i=1}^n (x_i-x_i')^2}$
for two vectors $\bm x=[x_1,\cdots,x_n]^T$ and $\bm x'=[x_1',\cdots,x_n']^T$,
in which each component is within the interval $[0,1]$.

Note that by our definition, $\varepsilon=1$ is a large perturbation.
For example, for two vectors from MNIST,
if their grey value difference is 10 at each component, then $\varepsilon=\sqrt{(\frac {10}{255})^2\cdot 28^2}\approx 1.1$.

For 3), we use the fooling ratio as the index to show the defensive capability.
Its definition is as follows. Given a set
of originally correctly classified samples, after adding
perturbation, some samples become misclassified.
The ratio of the number of misclassified sample to the total sample number
is called the fooling ratio.
Clearly, the lower the fooling ratio, the better the performance of the network.

For  4),
we use ten attacking methods for white-box and black-box attacks, respectively.
Most of them have been discussed in Section 2.
The white-box attack assumes that the attackers know all the details of the classifier, including the model and parameters, while the black-box attack assumes all the details are unknown and the adversarial examples are fed into the network.

\subsubsection{Datasets and classification methods used in the experiments}
In the experiments, three datasets, together with corresponding neural networks, are utilized.
These datasets and networks consist of
1) MNIST+ConvNet, 2) CIFAR-10+VGG, and 3) German Traffic Sign Recognition Benchmark (GTSRB)+ResNet.
The first two datasets are common benchmarks, and the third one is used as
a demonstration for real applications such as autonomous driving.

MNIST \cite{MNIST} is  equipped with a
simple CNN network, which is a variant of LeNet and called ConvNet here. ConvNet has two convolutional layers with batch normalization and one fully connected layer.
Before defensive training, this simple network can reach a classification accuracy of 99\%.

CIFAR-10 \cite{krizhevsky2009learning} is equipped with
a simplified VGG.
This VGG network adopts a variant of simplified 11 layers in order to fit the dataset size.
Before defensive training, this  model can reach a classification accuracy of around 90.5\%.

GTSRB \footnote{http://benchmark.ini.rub.de} \cite{Germany} consists of more than 50,000 images of
traffic signal in 43 classes. This dataset has  39,209 training images and 12,630
testing images. To facilitate the procedure, we use 39,200 training and 12,600 testing
images and rescale the image size to $32\times 32$. GTSRB is equipped with a simplified
ResNet with 14 layers to accelerate the training. Before defensive training,
the accuracy is around 98.5\%.

\subsection{Defending against white-box attacks}

We compare the fooling ratio under 1) various
values of the regularization coefficient, 2) various datasets and neural network models, and 3) various attacking methods.

\subsubsection{Fooling ratios with various values of the regularization coefficient}
We compare the capability for generating adversarial examples under various
values of the regularization
coefficient $\mu$, all other parameters are set as the same, i.e., dataset+DNN: MNIST+ConvNet, and the attacking method for generating adversarial examples: OSSA.
The purpose of this experiment is to show the effectiveness of our scheme when a suitable
value of $\mu>0$ is set.

We wish to observe
that fooling ratio becomes low, compared with the original network, while the accuracy for the testing data does not decrease using our scheme.

\begin{figure}
\centering
\includegraphics[width=0.5\textwidth]{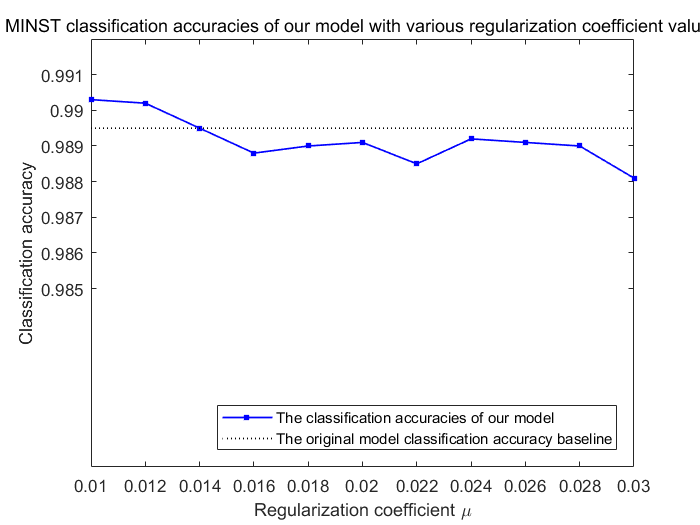}
\caption{Classification accuracies of the baseline and our scheme with various coefficient values
on MNIST under OSSA.}
\label{fig_accuracy}
\end{figure}

\begin{figure}
\centering
\includegraphics[width=0.5\textwidth]{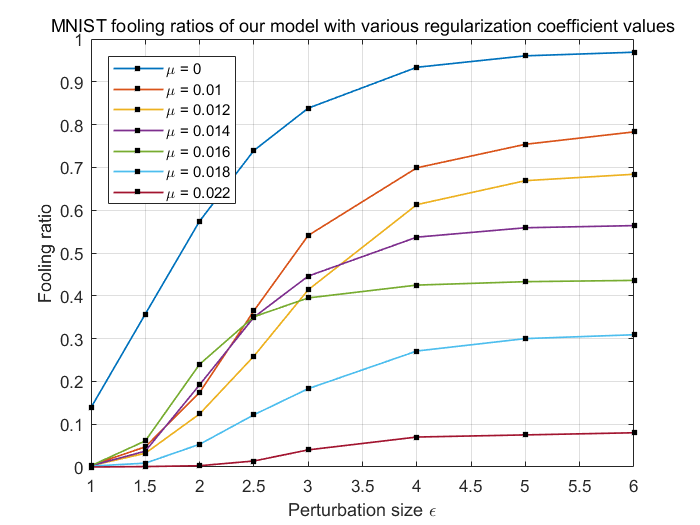}
\caption{The fooling ratio curves with various regularization coefficient values. All data are obtained on MNIST, using OSSA.}
\label{fig_ratios_lambda}
\end{figure}

Figures \ref{fig_accuracy} and \ref{fig_ratios_lambda} verify the above claim. Figure \ref{fig_accuracy} shows that, using our scheme,
the accuracy remains virtually unchanged, while
Figure \ref{fig_ratios_lambda} shows that the fooling ratio significantly decreases.

In Figure \ref{fig_accuracy}, the dotted line is the baseline classification accuracy of
the original network, i.e., the ConvNet without defensive training, on MNIST.
The dotted line is obtained by averaging the accuracies of several experiments.
Figure \ref{fig_accuracy} shows that the accuracy remains virtually unchanged
(fluctuated around 98.95\%) using our scheme.

Figure \ref{fig_ratios_lambda} shows that, under certain range of parameters,
with the increment
of the regularization coefficient $\mu$, the fooling ratio of the generated adversarial
examples decreases monotonically. When $\mu = 0.022$ and $\varepsilon=6$
($\varepsilon=6$ is a large distance, as we discussed previously)
the fooling ratio is suppressed significantly
small (less than 10\%), while the classification accuracy is still as high as 98.8\% in
Figure \ref{fig_accuracy}.
Since the fooling ratio reflects the defensive capability, low fooling ratio means that the defensive capability is satisfactory.

Figures \ref{fig_accuracy} and \ref{fig_ratios_lambda} demonstrate that our scheme obtains the defensive capability
without sacrificing the accuracy.

\subsubsection{Fooling ratios on various datasets and DNNs}

In this subsection, we compare the defensive capability for our scheme under various datasets+DNNs, on the same attacking algorithm OSSA. The purpose of this experiment is to evaluate our robustness on different datasets and DNNs.

\begin{figure}[!htbp]
\centering
\includegraphics[width=0.5\textwidth]{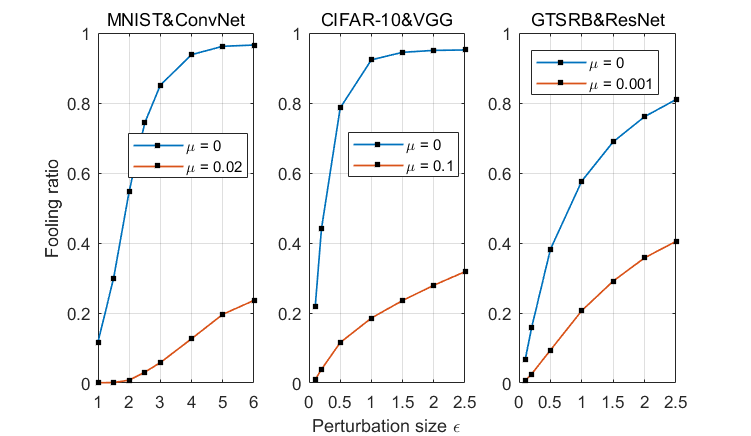}
\caption{The fooling ratio curves on three datasets with corresponding DNN models, with or without
adding $\mu$. All experimental data are generated by OSSA.}
\label{fig_three_datasets}
\end{figure}

Figure \ref{fig_three_datasets} shows the fooling ratio curves under three datasets and their
corresponding DNNs. It reveals that, for all cases, the fooling ratio decreases when a suitable $\mu>0$ is set.

\subsubsection{Fooling ratios on various attacking methods}

In this subsection, we compare the defensive capability of our scheme
under various attacks. The defensive scheme is set as $\mu=0.02$ for MNIST+ConvNet.
Three non-OSSA one-step attack methods,
namely, FGM \cite{kurakin2016adversarial}, OTCM \cite{kurakin2016adversarial} and FGSM,
are used for the experiment.
All three attacking methods are implemented by ourselves. Here OTCM
is the acronym for
One-step Target Class Method.

The purpose of this experiment is to verify the robustness on various attacking methods.

\begin{figure}[!htbp]
\centering
\includegraphics[width=0.5\textwidth]{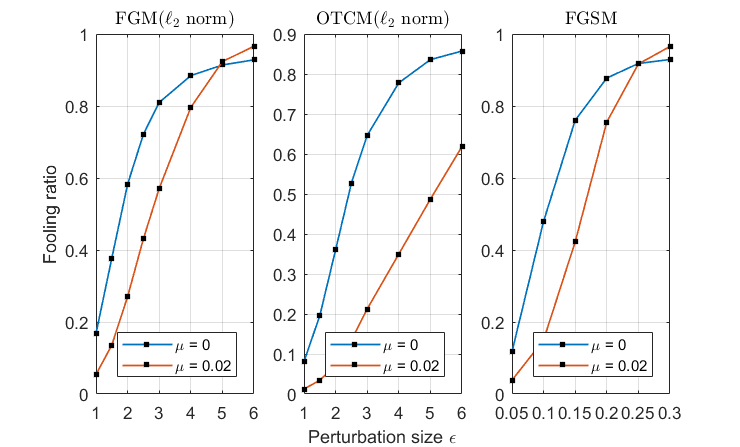}
\caption{The fooling ratio curves of three adversarial attacking methods on MNIST.
}
\label{fig_three_methods}
\end{figure}

Figure \ref{fig_three_methods} shows that the fooling ratios are low,
when the perturbation norm are small (which is the case we are interested in).
However, these fooling ratios are high compared with their counterparts in Figure \ref{fig_three_datasets}.
This is reasonable that our method is inspired by \cite{our_AAAI} where OSSA is proposed.
When $\varepsilon$ is large, e.g., $\varepsilon=6$, our scheme fails as the fooling ratio
is higher than 60\%.

\textbf{Defensive capability on more attacking methods using FoolBox}
To further evaluate the defensive capability  under more
attacks,
we adopt a toolbox called
FoolBox\footnote{https://foolbox.readthedocs.io/en/latest/index.html\#}
\cite{deepfool} which provides various popular attacking methods.

The toolbox aims to obtain the optimized adversarial examples,
i.e., the perturbation should be as small as possible when generating the adversarial examples.
So, for a given perturbation norm, it does not generate adversarial examples accordingly.
Therefore, we use the distance between the original input and the adversarial
example as a measurement for the degree of  difficulty for generating an adversarial example in DNNs.
The larger the distance, the more difficult the capability of generating adversarial examples,
i.e., the better the defensive capability.

Table \ref{table_1} lists the distances between the original input and the adversarial
example, for the original network ($\mu=0$) and our scheme ($\mu=0.024$),
under nine attacking methods, with MNIST.
In the experiments, we load the pretrained model and samples from MNIST as the input for the testing.
Then we use the method provided in FoolBox to generate a batch of adversarial examples,
and obtain the distance between every adversarial example and its original input. The second column
shows the norms used in the attacks.
Columns 3 and 4 show the mean distance between  the adversarial and original inputs.
The above mean distance can be used as a measure for the degree of difficulty on
generating adversarial examples, as the same parameters are used for generating adversarial
examples for two models. The larger the distance, the more difficult for generating adversarial
examples. Table \ref{table_1} shows that, for all nine methods, our scheme outperforms
as our mean distances are larger, and the distance
ratio is between 1:1.3 and 1:1.8.
This demonstrates the robustness of our method.

\begin{table}[!htbp]
   \caption{Comparison of distances between the original input and its adversarial example for two models using MNIST.}
  \centering
  \begin{tabular}{ccrr}
  \hline
  Method        &Norm       &{Mean Distance\ \ }         &Mean Distance\\
                &           &on Original Model       &on Our Scheme\\
  \hline
  FGSM          &$l_\infty$ &0.107                      &\textbf{0.170}\\
  FGM           &$l_2$      &0.00584                    &\textbf{0.00791}\\
  BIM           &$l_1$      &0.0210                     &\textbf{0.0273}\\
  BIM           &$l_2$      &0.00158                    &\textbf{0.00273}\\
  BIM           &$l_\infty$ &0.0726                     &\textbf{0.1070}\\
  DeepFool      &$l_2$      &0.00215                    &\textbf{0.00267}\\
  CW            &$l_2$      &0.00145                    &\textbf{0.00204}\\
  JSMA          &$l_0$      &28.26                      &\textbf{48.19}\\
  Random PGD    &$l_\infty$ &0.0730                     &\textbf{0.1100}\\
  \hline
  \end{tabular}
  \label{table_1}
\end{table}

\subsection{Defending against black-box attacks}

Our scheme is further tested for defending against black-box
attacks.

Table \ref{table_2} lists the results on MNIST+ConvNet.
All adversarial examples are generated by FoolBox, except that OSSA in the last row
is generated by ourselves
with the parameter $\varepsilon=1.0$.
Specifically, given a sample, we use a method provided in the FoolBox and generate an
adversarial example for the Network $A$. This sample is adversarial for the original network $A$,
we test whether it is still adversarial for Network $B$. Similarly, we can also generate an
adversarial example for Network $B$ and see whether it is adversarial in Network $A$.
We wish to see, given a batch of samples, using the above operation, samples generated from
the original network still have high accuracy (i.e., low fooling ratio) by our scheme, while
adversarial examples of our scheme will have a low accuracy for the original network, as we assume
our scheme has a higher defensive capability than the original network.

In the table,
Columns 2 and 3 show the above
cross model classification accuracies.
The value (97.47\%) in row 1 and column 2 means that, for example,
we first use  FGSM from FoolBox to attack the original network and generate
adversarial examples, and input these adversarial examples to our scheme,
then in our scheme
the accuracy is 97.47\%.
This means that the accuracy raises from 0\% of the original
network to 97.47\% of our scheme.

Note that there is no value at the last row and last column. The reason is that
our scheme can defend against OSSA significantly and no adversarial examples can be
generated for $\varepsilon = 1.0$.
It seems that the accuracies in column 2  are higher than expected. We analyze the reason as
that the adversarial examples generated by FoolBox are optimal
with respect to perturbation, which means that
it may not be optimal for a defensive model and therefore its defense is relatively easy.
In general, the classification accuracy in column 3 is lower than that in column 2,
as we have analyzed. Thus we can conclude that
our model behaves reasonably acceptable for black-box attacks, although the difference
of accuracies is not sufficiently large.
It might be combined  with other defense techniques to achieve better performance.

\begin{table}[!htbp]
  \caption{Cross model classification accuracies on MNIST.}
  \label{table_2}
  \centering
  \begin{tabular}{ccc}
  \hline
                            &Generated from       &Generated from \\
      Attacking             &original network     &our scheme\\
      methods               &Tested on          &Tested on \\
                            &our scheme          &original network\\
  \hline
  FGSM ($l_\infty$)         &97.47\%            &54.33\%\\
  FGM ($l_2$)               &96.74\%            &78.18\%\\
  BIM ($l_1$)               &98.79\%            &93.12\% \\
  BIM ($l_2$)               &98.79\%            &93.72\%\\
  BIM ($l_\infty$)          &98.89\%            &89.07\%\\
  DeepFool ($l_2$)          &98.58\%            &93.64\%\\
  CW ($l_2$)                &98.79\%            &97.47\%\\
  JSMA ($l_0$)              &96.09\%            &90.45\%\\
  Random PGD ($l_\infty$)   &98.99\%            &89.57\%\\
  OSSA ($l_2$)              &92.24\%            & --\\
  \hline
  \end{tabular}

  \label{table_2}
\end{table}

\section{Conclusion}
\label{conclusion}
We have proposed a defensive scheme for
adversarial attacks by modifying the loss function
of DNNs to control the largest eigenvalue of the FIM.
Our scheme stems from the assumption that the
vulnerability of DNNs is due to the
the large value of the largest eigenvalue of FIM.
Elaborated by experimental results on
typical DNNs with various datasets,
our scheme demonstrates its capability of defending against adversarial attacks.

Our contributions can be summarized as:
1) Our scheme is a simple yet effective regularization method. Compared with other
adversarial training methods, it only needs to retrain once;
2) our scheme has an explainable property compared with LSR;
and 3) the effectiveness is demonstrated by various DNNs on common datasets.

Future work can be focused on using the theory of simplex to further improve
the loss function, as the density vectors are on a simplex.
\bibliographystyle{IEEEbib}
\bibliography{refs_arXiv}

\end{document}